\documentclass{article}

\usepackage{times}
\usepackage{microtype}
\usepackage{graphicx}
\usepackage{subfigure}
\usepackage{paralist}
\usepackage{tikz}
\usetikzlibrary{positioning,decorations.pathreplacing,shapes}

\usepackage{amsmath}
\usepackage{amssymb}
\usepackage{bm}

\usepackage{tabularx}
\usepackage{booktabs}
\usepackage{multirow}

\usepackage{mysymbols}

\usepackage{import}

\usepackage{hyperref}

\newcommand{\xb}{\mathbf{x}}

\newcommand{\yb}{\mathbf{y}}

\usepackage[accepted]{icml2018}

\icmltitlerunning{Learning Multi-modal Mappings from Sparse Annotations}

\begin{document}

\twocolumn[
\icmltitle{Learn from Your Neighbor: \\
Learning Multi-modal Mappings from Sparse Annotations}




\begin{icmlauthorlist}
\icmlauthor{Ashwin Kalyan}{gt}
\icmlauthor{Stefan Lee}{gt}
\icmlauthor{Anitha Kannan}{curai}
\icmlauthor{Dhruv Batra}{gt,fair}
\end{icmlauthorlist}

\icmlaffiliation{gt}{Georgia Tech}
\icmlaffiliation{fair}{Facebook AI Research}
\icmlaffiliation{curai}{Curai}

\icmlcorrespondingauthor{Ashwin Kalyan}{ashwinkv@gatech.edu}


\vskip 0.3in
]



\printAffiliationsAndNotice{} 
\begin{abstract}
Many structured prediction problems (particularly in vision and language domains) are ambiguous, with multiple outputs being `correct' for an input -- \eg there are many ways of describing an image, multiple ways of translating a sentence; 
however, exhaustively annotating the applicability of all possible outputs is intractable due to exponentially large output spaces (\eg all English sentences). 
In practice, these problems are cast as multi-class prediction, with the likelihood of only a sparse set of annotations being maximized -- unfortunately penalizing 
for placing beliefs on plausible but unannotated outputs. 
We make and test the following hypothesis -- for a given input, the annotations of its neighbors may serve as an additional supervisory signal. 
Specifically, we propose an objective that \emph{transfers} supervision from neighboring examples. 
We first study the properties of our developed method in a controlled toy setup before reporting results on 
multi-label classification and two image-grounded sequence modeling tasks -- captioning and question generation. 
We evaluate using standard task-specific metrics and measures of output diversity, 
finding consistent improvements over standard maximum likelihood training and other baselines.
\end{abstract}

\section{Introduction} \label{sec:intro}
\begin{figure}[t]
\centering    \includegraphics[width=0.9\columnwidth]{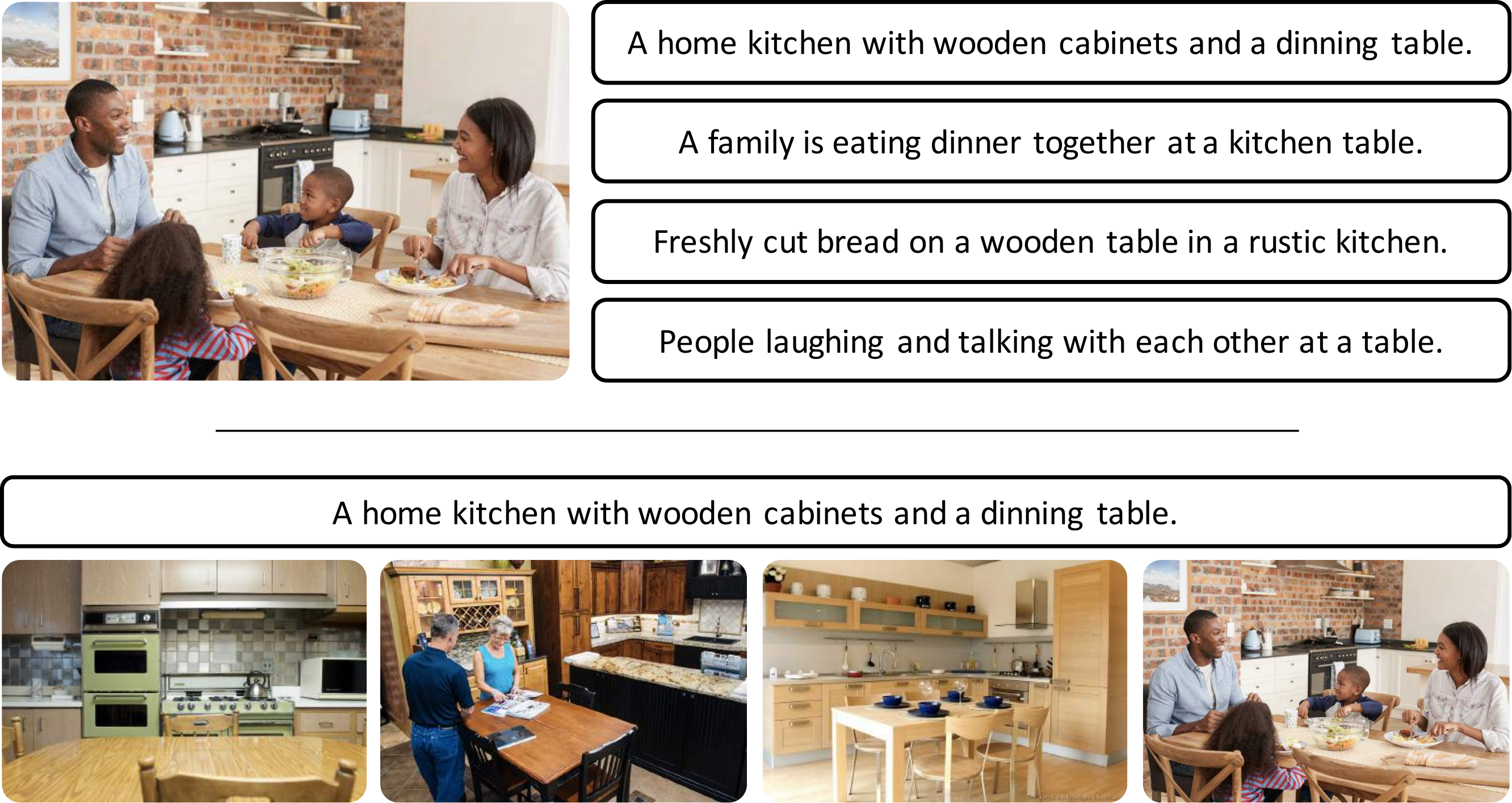}%
\caption{Many tasks exhibit many-to-many relationships between inputs and outputs.
    Taking image captioning as an example, a single image can be described with multiple captions (top) and likewise a single caption can accurately describe multiple images (bottom).
In this work, we leverage these relationships in the data to learn multi-modal output mappings from sparse annotations.}
    \label{fig:cover}
\end{figure}
In many real-world tasks, a single input is associated with multiple correct outputs. 
For instance, as shown in \figref{fig:cover}, multiple captions can accurately describe an image.
For tasks with small output spaces, it may be practical to treat the problem under a multi-label formulation -- learning to predict the correctness of each possible output based on exhaustive human annotations. 
However, for structured prediction tasks, the output space is exponentially large (\eg the space of all English sentences) such that collecting exhuastive annotations is intractable even for a single input. 
Instead, sparse annotations are obtained by collecting human responses -- leaving the correctness of a vast majority of possible outputs uncertain.
\\ \\
This problem of \emph{multi-label classification with missing labels} has been addressed in prior work by either imposing structure on the label space such as known label taxonomies \cite{verma_bmvc13, deng_eccv14}, or by imposing constraints on the model parameters \cite{yu_icml14} or the posterior distributions \cite{lin_nipsw14} to effectively compress the label space. 
However, these approaches do not scale to the exponentially large label spaces often seen in structured prediction tasks like sequence-modeling (\eg $|\calV|^T$ length--$T$ sentences in captioning where $\calV$ is the vocabulary).
As a consequence, such tasks are often cast as multi-class problems with parameters learned to maximize the likelihood of a sparse set of human annotations (\ie Maximum Likelihood training) -- implicitly enforcing the unreasonable assumption that all outputs that are not annotated must be incorrect. 

Much contemporary research has been invested in the more expensive yet viable option of curating massive datasets that leads to better estimation of the true multi-modal mapping with increasing dataset size.
For instance, progress in captioning has been largely propelled by the massive COCO \cite{coco} dataset containing ${\sim}330K$ images with $5$ human-provided captions per image. 
However, as evidenced by the impoverished, generic captions generated by models trained on this dataset \cite{vijayakumar_arxiv16, dai_arxiv17}, even such large-scale efforts fall short -- capturing only a small fraction of all possible outputs. 


\textbf{Overview and Contributions.} In this work, we propose a simple approach that enables models to place beliefs on multiple plausible 
outputs while training only on sparse set of annotations, or in the extreme case only a single annotation per input on tasks 
where there are many many possible correct outputs. Essentially, our goal is to learn to produce multi-modal outputs 
from `uni-modal' annotations. 
The key inductive bias in our approach is the following -- for a given input, 
the annotations of its neighbors may serve as an additional supervisory signal. 
\figref{fig:cover} (bottom) demonstrates this intuition for captioning with the caption accurately describing all four depicted scenes. 
Based on this insight, we propose a novel objective that treats outputs of neighboring inputs to be applicable to the given input to an extent determined by the similarity in the input space.  
This objective allows us to \emph{transfer} annotations from neighboring examples to provide additional supervision 
and so contribute towards recovering the underlying multi-modal mapping. 


In order to analyze our approach in a tractable domain, we perform a number of multi-label classification experiments with missing labels. 
First, we evaluate in a toy setting where the data generating distribution is known and find that our method is able to better estimate the true distribution as compared to standard cross-entropy training. 
We also study multi-label prediction on two real-world datasets -- CUB-200 \cite{wah2011caltech} and Animals with Attributes (AWA) \cite{xian2017zero} -- by sub-sampling attribute annotations.
As in the toy setting, we see improvements over baseline methods. 
Finally, we apply our method to two established image-grounded language generation tasks -- image captioning and question generation -- which are both sequential prediction tasks with exponentially large label spaces. 
We evaluate using both standard task-specific metrics for the generated language and criteria that assess the multi-modal nature of the produced outputs. 
We find consistent improvements over baseline methods on these challenging tasks.

\section{Approach}\label{sec:app}
\label{sec:problem}
\noindent We first establish the notation and succinctly summarize the learning problem before explaining the proposed approach.

Consider a multi-label prediction setting where the goal is to learn a \emph{one-to-many} relationship $f{:}\calX\rightarrow2^\calY/{\emptyset}$ that maps a given input $\xb{\in}\calX$ to a set of valid outputs, a subset of all possible outputs $\calY$. 
In our setting, obtaining annotations for each element in $\calY$ is intractable even for a single instance and instead only a sparse set of \emph{positive} annotations are available.
Specifically, we assume access only to a dataset of the form $\calD = \{(\xb_m, \{y_{m,1}, \dots, y_{m,k}\})\}_{m=1}^M$ where $\xb_m$ is the input and $\{y_{m,1}, \dots, y_{m,k}\}$ is a sparse set of labellings with $k << |f(\xb_m)|$. 
In practice, $k$ may vary for each example and often, $k=1$ \ie only one annotation is available. 

The observed dataset $\calD$ can be thought of as being produced by a stochastic function $g$ that selects $k$ labels from $f(\xb)$, the set of all applicable labels for $\xb$. In practice, a collection of human annotators often play the role of $g$, generating a small set of possible outputs for each $\xb$ (\eg each providing a single image description in captioning). 
We therefore summarize the overall learning problem as -- \emph{how can we estimate the true multi-modal input-output relationship $f$ while only observing sparse samples from $g$?}

\begin{figure}[b]
    \centering
    \includegraphics[width=\columnwidth, clip=true, trim=10px 20px 10px 10px]{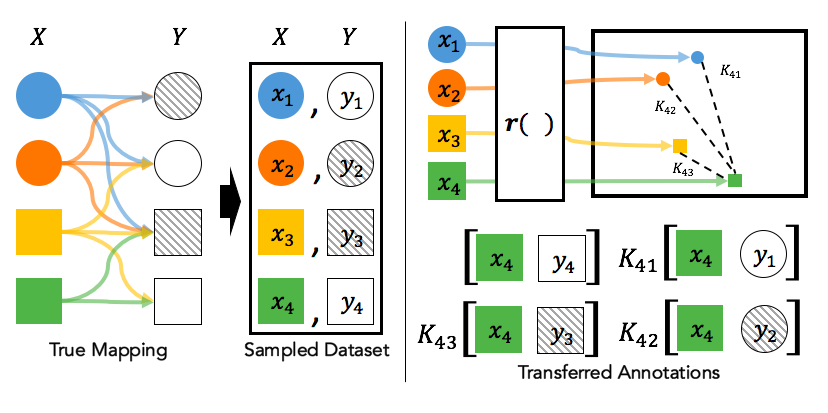}\\[-5pt]
\caption{Given a dataset of sparse annotations sampled from a true multi-modal input-output mapping (left), our approach leverages a learned similarity space to perform a soft-transfer of annotations between semantically related inputs (right) -- effectively recovering the underlying multi-modal mapping from few samples.}
    \label{fig:approach}
\end{figure}%

\subsection{Enforcing Adaptive Neighborhood Structures} \label{subsec:objective}

At a high-level, our approach has two key components -- \\
1) a mechanism that allows us to use outputs of neighboring inputs to provide additional supervision  and 
2) an appropriate measure of semantic relatedness to define the neighborhood.
We now explain both these aspects in detail.


\textbf{Learning from Neighbors.} The predictions $\tilde{y}_m$ outputted by the model for each input $\xb_m$ are evaluated using a loss function $\ell:\calY\times\calY\rightarrow\RR$ and the standard objective is to reduce the empirical risk on the training set:
\begin{equation}
\frac{1}{Mk}\sum_{m=1}^M\sum_{k}\ell\left(\tilde{y}_m, ~y_{m,k}\right)
\label{eq:erm}
\end{equation}
Due to the presence of multiple annotations, the summations cover both examples ($m$) and their annotations ($k$). 

Let us begin by assuming access to a function $r:\calX\rightarrow\RR^d$, 
a potentially non-linear transformation that maps inputs to a space where distances correspond to semantic relatedness. 
Let $\calK_{ij}$ be the \emph{similarity} between $\xb_i$ and $\xb_j$ in this semantic space.
Note that all distances and similarities are computed in this semantic space unless otherwise mentioned.
Equipped with this semantic space, we now define the neighborhood $\calN(\xb)$ of a data point $\xb$ -- specifically, let $\calN(\xb)$ to be the set of indices of the $N$-nearest neighbors of $\xb$.
Recall that we wish to incorporate the key inductive bias that outputs of semantically similar inputs can be potentially `correct' for a given input.
Thus, we can now write a regularized objective that encourages the model to place beliefs on multiple outputs as:
%
\vspace{-5pt}
\begin{equation}
    \underbrace{\ell(\tilde{y}_i, y_{i,k})}_{\text{loss \wrt own label}} 
\hspace{-5pt}
    + \underbrace{\frac{\lambda}{\tiny \abs{\calN(\xb_i)}}}_\text{normalization}
    \sum_{j\in\calN(x_i)} 
\hspace{-25pt}    
    \overbrace{\calK_{ij}}^\text{similarity to neighbor}
\hspace{-35pt}
    \underbrace{\ell(\tilde{y_i},y_j)}_{\text{loss \wrt neighbor's label}} 
\label{eq:our}
\end{equation}
%
where $\lambda$ is a hyper-parameter that controls the importance of additional supervision.
The weighting of the additional supervision using the similarity $\calK_{ij}$ can be thought of as accounting for the \emph{uncertainty} in the applicability of neighboring output $y_j$ to the input $\xb_i$, due to the lack of its annotation.  
In this work, we set $\displaystyle \calK_{ij} = \max\Big\{0, \cos\big(r(\xb_i), r(\xb_j) \big)\Big\}$ where $\cos$ is the cosine similarity. 

\textbf{Connections to label smoothing.} Unlike maximum likelihood training that penalizes unannotated predictions, our objective encourages the model to place beliefs on outputs of neighboring inputs apart from its own annotatation. 
In a simple $\calC$-way classification setting, it is easy to see that this corresponds to label smoothing with class $c\in\calC$ assigned a mass proportional to ${\sum_{j\in\calN(\xb), g(\xb_j)=c} \calK_{ij}}$.
Thus, our loss redistributes mass in a systematic, input-aware fashion unlike \citet{szegedy_arxiv15} or \citet{pereyra_arxiv17} that uniformly increase the uncertainty in the predictions. 

\noindent \textbf{Learning the semantic space.} \label{sec: refine}
As mentioned before, computation of the neighborhood $\calN(\xb)$ and the similarities $\calK_{ij}$ assumes access to a projection $r(\cdot)$ that maps inputs to a semantic space.
Unless strong priors exist like known taxonomies exist in the input space, there is no obvious choice for this projection. 
As such, we propose learning it alongside the task -- specifically, we initialize $r(\cdot)$ with some domain specific neural network (\eg pre-final layer of a CNN on ImageNet \cite{imagenet} for natural images) and then finetune it jointly with the model. 

In practice, if the network for learning $r(\cdot)$ has sufficient capacity, it can project all points in the dataset to a unique dimension of its own \st $\calK_{ij} = 0 \ \forall i, \forall j, i\neq j$, reducing our objective to MLE, \eqref{eq:erm}.
Further, looking at the derivative of the objective \wrt $\calK_{ij}$ 
\[ \ddel{\calL_i}{\calK_{ij}} = \frac{\lambda\ell(\tilde{y_i},y_j)}{{\abs{\calN(\xb)}}}  \]
we can see that the objective constantly works towards reducing the similarity $\calK_{ij}$ as $\ell(\cdot,\cdot)$ is always non-negative.
To constrain the network from pushing similar data points apart, we regularize by penalizing the model for deviating too much from the initial structure as --
\begin{equation}
    \frac{\mu}{\abs{\calN(\xb_i)}} \sum_{j\in\calN(\xb_i)}(\calK_{ij}-1)^2
\label{eq:reg}
\end{equation}
\noindent where $\mu$ is a hyper-parameter that controls the strength of this penalty. 
Note that this penalty implies that our initial choice for $r(\cdot)$ is already reasonable and only requires minor adjustments for it to be task-specific.

\subsection{Generalization to Sequence Prediction}
\label{subsec:fac_loss}
\noindent 

\begin{figure}[t]
    \centering
  \includegraphics[height=5cm, clip=true, trim=80pt 20pt 170pt 0pt]{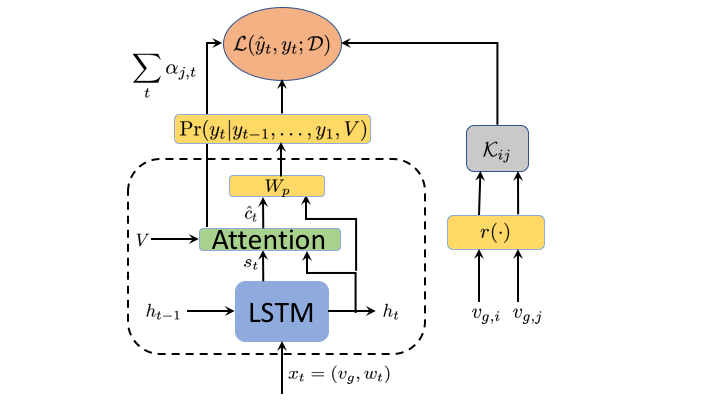}\\[-5pt]
    \caption{An diagram of our approach for sequential prediction task. Only the relevant segments (via a soft attention mechanism) from the neighbors output are used. See Sec.~\ref{subsec:fac_loss} for details.}
    \label{fig:model}
\end{figure}

Consider sequential output tasks where an input $\xb$ is mapped to a sequence $\yb = \{y_1, \dots, y_T\}$.
The standard objective for seqeunce modeling is to maximize $\log$-likelihood of the output token at time $t$ given previous tokens and the input as $\sum_t \log\prob(y_t | y_{t-1}, \dots y_1, \xb)$.
We can trivially extend our objective in Eq. \eqref{eq:our} by weighing each term inside the summation with $\calK_{ij}$.
However, for grounded sequence generation tasks like captioning, it is often the case that only a portion of the neighbor's output is applicable to a given input; for instance, a pair of images may both contain a dog, but only one also has a cat. 
In such cases, only specific phrases or words may be reasonably borrowed between images (\ie ``big dog''). 
To incorporate this notion of `partial' supervision, we extend our objective to leverage attentional models like that of \citet{lu_cvpr17}.
%
\\ \\
We now briefly explain this attentional model and refer the reader to \citet{lu_cvpr17} for a more detailed discussion. 
Consider a set of visual features $V=\{v_1, \dots, v_k\}$ that each encode different regions of the image and a global feature $v_g$ given by their average.
As shown in \figref{fig:model}, the model takes in this global image feature, $w_t$ an embedding of the previous word $y_t$ and the spatial features $V$ to compute the attention vector, $\bm{\alpha}_t\in\RR^{k+1}$ that weighs the importance of each of the spatial regions and the history encoding $h_t$ to compute the posterior $\log\prob(y_t | y_{t-1}, \dots, \xb)$.
Unlike standard attention-based architectures \cite{xu_icml15} that only attend to image regions, \citet{lu_cvpr17} extend it to incorporate both image and language components. 
Interpreting the sum of visual attention weights, denoted by $\alpha_t\in[0,1]$ as the importance of the image for the generation of the next word allows us to incorporate the notion of only copying partial sequences by weighing each term in the factored $\log$-likelihood by this visual importance weight.
Specifically, let $\alpha_{j,t}$ be the \emph{visual} importance of the neighbor $\xb_j$ for predicting the word $y_{j,t}$ given $\xb_i$.  
Then, the second term in \eqref{eq:our} can be modified as:
\begin{equation}
    -\frac{\lambda}{\calN(\xb)} \sum_{j\in\calN(\xb)} \calK_{i,j} \sum_{t\in[T]} \alpha_{j,t} \log\prob(y_{j,t} | y_{j,t-1}, \dots, y_1, \xb_i)
\end{equation}
Likewise, the regularization constraint from \eqref{eq:reg} is updated to ensure that $\alpha$'s do not go to zero:
\begin{equation}
\frac{\mu}{T\calN(\xb)} \sum_{j\in\calN(\xb)} \calK_{ij} \left(\sum_{t\in[T]} \alpha_{j,t} - 1\right)^2.
\end{equation}
\vspace{-10pt}
\subsection{Implementation Details.}
\label{sec:implementation_details}
\noindent We now discuss some subtle but important implementation details training models using our objective. 
\\ \\ 
\noindent Each mini-batch consists of $B$ examples and their corresponding $N$ neighbors (including itself) -- resulting in a total of $B\times N$ samples that need to be processed for an \emph{effective} batch-size of $B$. 
The procedure for sampling neighbors has to be ``aware'' of the constantly changing representation space $r(\cdot)$ and so, is also updated in tandem.
We call this \emph{adaptive} updation of the neighborhood of each image. 
As it is expensive to compute similarities and sample these mini-batches, we make the following practical choices --  
First, the parameters of $r(\cdot)$ are updated using a \emph{much} smaller learning rate ($10\times$) compared to the model itself.
Second, the similarities between data points and thus the neighborhood is only updated every few iterations. 
\\ \\ 
Finally, our method is fairly robust to the setting of $\lambda$ and $\mu$, the repulsive and attractive terms in our objective. 
In general, as the number of neighbors participating in the objective increases smaller values of $\lambda$ suffice. 
Further, large values of $\mu$ enforce the neighborhood to strongly respect the initial structure and so, depending on the quality of the initialization of this representation space, $\mu$ can be varied. 
A more detailed analysis of the these hyper-parameters is provided in the supplement.

\section{Related Work} \label{sec:rel}
\noindent\textbf{Multi-label Classification with missing labels.}
\cite{verma_bmvc13} extend the work of \citet{bucak_cvpr11} by incorporating taxonomy of the label space into their cost-senstive ranking formulation. 
Extending these methods to sequential outputs is challenging as there is no documented or natural way of constructing such a taxonomy for say, English sentences. 
Instead, our approach jointly learns dependencies between data points and does not require similarity or taxonomy information as input.
Further, \citet{yu_icml14} propose imposing a low-rank constraint on the weights to be able to capture correlation between labels. 
However, such norm-based regularization schemes have been showed to be ineffective for deep networks \cite{zhang_arxiv16}. 
In a similar vein, \citet{lin_nipsw14} constrain the posterior to be of low rank. 
However, such an approach is not feasible when the output is a sequence. 
\\ \\
\noindent\textbf{Semi-supervised learning.}  
The use of homophily for semi-supervised learning has been well-studied (c.f. \citet{zhu2005semi,zhu2003semi} for a survey). 
A dominant approach is the use of relationship graphs as regularization -- these methods assume access to a relationship graph (that is not available in our setting) as part of the input, or that it can be easily gleaned  through measuring similarity in the input space; the underlying assumption being that the network structure is independent of the labels given the input. 
\citet{weston2012deep} extends this line of work by embedding inputs using a deep neural network. \citet{perozzi2014deepwalk} and \citet{yang2016revisiting} extends this to infer graph context to aid in classification. 
In general, the goal of this line of work is to enable a consensus in the label assignment of unlabeled examples, by leveraging the neighborhood structure. 
However, in our setting all examples are labeled, albeit incompletely (missing labels). 
The  central hypothesis of our work is that there exists an embedding space in which neighborhood structure is evident where neighboring data points can be used to supervise the learning of a multi-modal output space. Thus, the model jointly optimizes for both -- uncovering this underlying semantic structure in the data while also, learning a multi-modal input-output mapping. 
\\ \\
\noindent\textbf{Entropy Regularization and Label smoothing}
\citet{pereyra_arxiv17} propose regularizing the model with a negative entropy term and in a similar vein, \citet{szegedy_arxiv15} propose a simple label smoothing strategy where an arbitrarily small probability mass is re-distributed to classes other than the observed ground-truth uniformly. 
Further, \citet{chorowski_arxiv16} draw from \citet{szegedy_arxiv15} and propose a neighborhood smoothing scheme for a $n$-gram language model where probability mass is distributed based on observed $n$-grams in the dataset and not uniformly like the previous work. 
In contrast, the goal of our formulation is to not just increase or decrease the prediction entropy but to \emph{shape} the conditional $\prob(y|\xb)$ to reflect multi-modal input-output mappings by placing meaningful beliefs on the output space.
In that spirit, our work shares motivation with the line of work on producing diverse structured outputs -- 
\cite{batra_eccv12, rivera_nips12, prasad_nips14, guzman2014efficiently, lee_nips16}.

\noindent\textbf{Non-MLE based Image-captioning.}
Interestingly, \citet{dai_arxiv17, shetty_arxiv17} obtain diverse outputs (relative to MLE) using adversarial training  without making any explicit assumptions about the multi-modal nature of the task.
As \citet{shetty_arxiv17} requires a multi-modal dataset, we instead compare to \citet{dai_arxiv17} and show that we outperform when only having access to limited uni-modal data.
Similarly, \citet{jain_arxiv17} use variational auto-encoders for the task of producing visually grounded questions and report diverse outputs. 
However, as observed with adversarial training, without multiple output annotations the latent variable does not contribute in capturing the multi-modal output space.
Finally, we also compare to \citet{rennie_arxiv16} that directly optimizes for the task-specific metric (like CIDEr~\cite{vedantam_cvpr15} or SPICE \cite{anderson_eccv16} for image-captioning) using policy gradients and show that we outperform their approach in our setting.

\textbf{Nearest-neighbor based captioning.} While \citet{devlin2015exploring} explore a nearest-neighbor based approach to captioning, \citet{chen2017reference} build on it to propose a modified objective that weighs each word based on its occurrence in nearest neighbor images. 
Similarly, \citet{mun2017text} propose an attention scheme that factors in the \emph{consensus} caption. 
Unlike our approach, both these methods push the model towards producing more generic descriptions that are applicable to multiple similar images.

\noindent\textbf{Similarity based on outputs.}
\citet{inan_arxiv16} propose a re-use of the word-embeddings by augmenting the cross-entropy loss with a KL divergence term between the predictions and the normalized vector of the dot-products between the target-word embedding and the entire vocabulary. 
This term encourages the model to place belief on completions that are not necessarily observed in the dataset. 
While the high-level goals of both our objective and this work are similar, this approach relies only on distances in the output space and does not factor that the language generated can be conditioned on inputs in a different perceptual modality.

\section{Experiments} \label{sec:exp}
\noindent 
We first explore the properties of our objective in a controlled toy setting and evaluate the performance \wrt the true data distribution. 
 For completeness, we then show results on the multi-label with missing labels task on standard multi-label attribute datasets and finally, discuss the performance of our method on two visually-grounded language generation tasks -- captioning and question generation.

\subsection{Synthetic Experiments} \label{sec:toy_exp}
\newcommand{\rulesep}{\unskip\ ~~\vrule~~}

\begin{figure*}[t]

  \hspace{-90pt}\includegraphics[width=0.15\textwidth, trim=0px -2px 0pt 0pt]{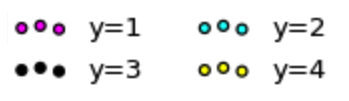}%
  \hspace{35pt}\includegraphics[width=0.35\textwidth]{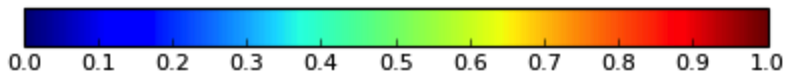}\\[-15pt]
  	\centering
	\subfigure[Synthetic Dataset]{
    	\label{fig:data_dist}
      \includegraphics[height=3.3cm, clip=true, trim=28px 25px 25px 25px]{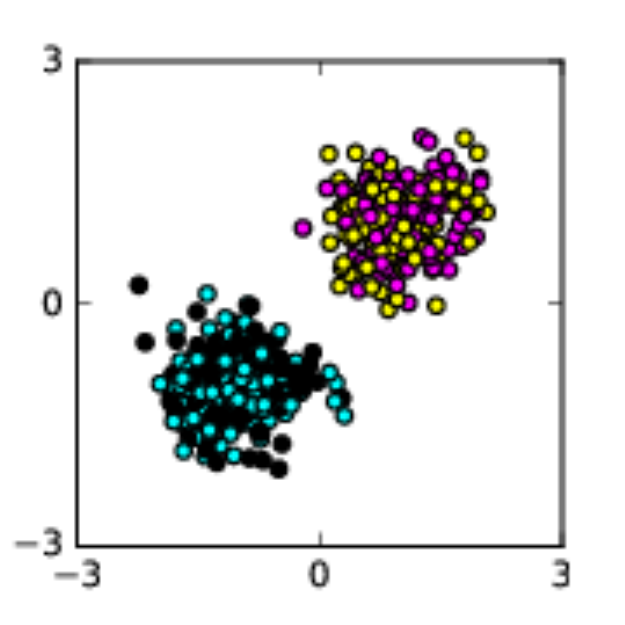}}\rulesep
	  \subfigure[$P(y{=}1|x)$: CE+L2]{
      \label{fig:ce_pred}
      \includegraphics[height=3.3cm, clip=true, trim=28px 25px 25px 25px]{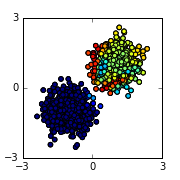}}
      \subfigure[$P(y{=}1|x)$: Ours]{
      \label{fig:our_pred}
        \includegraphics[height=3.3cm, clip=true, trim=28px 25px 25px 25px]{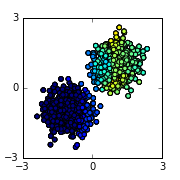}}\rulesep
      \subfigure[Sample Efficiency]{
      \label{fig:sample_plot}
      \includegraphics[height=3.9cm, clip=true, trim=30px 20px 50px 20px]{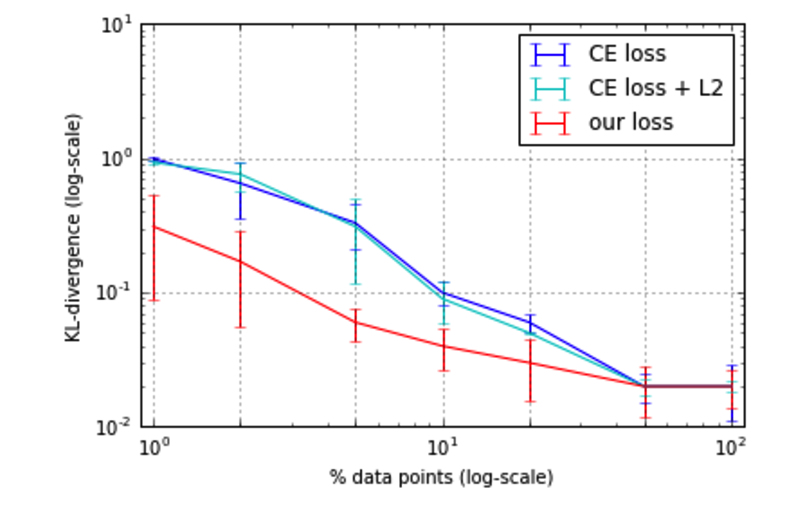}
      }\\[-6pt]
\caption{(a) Our toy experiment uses synthetic data with uniform label mixing within each cluster. 
    (b) We find that cross-entropy (CE) training (even with L2 regularization) results in overfitting -- for instance, some regions of the rightmost cluster are very confidently predicted as class 1 despite the true distribution being unbiased. (c) In contrast, 
our approach accurately predicts equal likelihood within clusters. 
 (d) Compared to training with CE loss, our approach accurately matches the true distribution as seen 
 by the significantly lower KL-divergence values \wrt the true-distribution while utilizing an order of magnitude fewer samples.}
\label{fig:toy}
\vspace{-10pt}
\end{figure*}

We consider a 4-label classification problem, where the dataset $\mathcal{D} =\{(\xb, y)\}$ is generated according to the graphical model shown below.

\hspace{20pt}\resizebox{!}{0.46\columnwidth}{
        \begin{tikzpicture}[
            node distance=0.25cm and 0cm,
            mynode/.style={draw,circle,thick,text width=1cm,align=center},
            scale=0.5
          ]
          \node[mynode] (z) {\huge$z$};
          \node[mynode,below right=0.5cm and 3cm of z,] (y) {\huge$y$};
          \node[mynode,above right=0.5cm and 3cm of z] (x) {\huge$x$};
          \path (z) edge[-latex] (x)
          (z) edge[-latex,thick] (y);
          (y) edge[latex-,thick] (x);
          \node[left=0.1cm of z]
          {
            $z\sim Bern(p_c)$
          };
          \node[above left=-20pt and 0pt of x]
          {
            $Pr(x|z) = \\
            \begin{cases}
            \mathcal{N}(\boldsymbol{\mu}_1,\Sigma_1)\text{, if }z{=}0 \\
            \mathcal{N}(\boldsymbol{\mu}_2,\Sigma_2)\text{, if }z{=}1
            \end{cases}
            $
          };
          \node[below left=-20pt and 0pt of y]
          {
            $Pr(y|z) =
            \begin{cases}
                [1{-}p_1, p_1, 0, 0]\text{, if }z{=}0 \\
                [0,0,1{-}p_2, p_2]\text{, if }z{=}1
            \end{cases}
            $
          };
          \end{tikzpicture}}

          Specifically, each data point $\xb$ is sampled from one of two Gaussians -- $\mathcal{N}(\boldsymbol{\mu}_1,\Sigma_1)$ or $\mathcal{N}(\boldsymbol{\mu}_1,\Sigma_2)$ depending on the state of the latent variable $z$.
          Each Gaussian is associated with two of the four labels $\left(z_0\rightarrow\{y_1, y_2\}\text{ and }z_1\rightarrow\{y_3,y_4\}\right)$.
However, for each data point $\xb$ we observe only one of the two possible labels. 
Using terminology from \secref{sec:problem}, the true multi-modal mapping $f$ maps the input in each cluster to two labels; however, we only observe one due to a stochastic label sampling function $g$. 
\figref{fig:data_dist} shows a plot of a dataset generated through this process (means of $(-1,-1)$ and $(1,1)$ with a diagonal variance of 0.2). 
To simulate the data-sparse regimes typical of real-world tasks, we transform the data to a much higher dimensional space (\eg $2^{13}$ in our experiments)  through a randomly initialized deep neural network with ReLU activations that doubles the input dimensionality.
For a trained classifier to perform well, it needs to discover the 2D data manifold that reveals the underlying neighborhood structure on which labels are based; and not simply overfit to local statistics in the high dimensional space.
Since the underlying data generation process is known, we can examine the hypothesis using KL divergence between the true posterior and the predicted distribution of trained models. 

\textbf{Implementation Details.} In our experiments, we use a dataset of size 2048; 512 for training and the rest to for evaluation.
We use a two-layered neural network with 32 neurons in each layer and train it via SGD with a learning rate of $4e{-}5$ and a momentum of 0.9. 
Further, we also compare with training using cross-entropy (CE) in conjunction with simple L2 regularization to show that our objective goes beyond such simple regularization schemes. 

\noindent In these experiments, we find evidence that our method 
\begin{compactenum}[\hspace{0pt} 1.]
\item \textbf{Induces smoothness in the conditional distributions. } \figref{fig:toy} shows a setting where both CE and its L2 regularized version obtain similar test losses but differ drastically in the label assignments compared to our objective. 
Specifically, \figref{fig:ce_pred} shows the conditional probability $P(y=1 | x)$ for test points from a CE+L2 trained model. Even for the L2 regularized model, there is significant overfitting with some regions of the rightmost Gaussian confidently predicted as class 1. In contrast, our approach shown in \figref{fig:our_pred} results in near uniform probability between classes 1 and 2 within the cluster. 
\item \textbf{Acts as a regularizer.} Since our objective enforces that neighboring data points have similar output distributions, over-fitting by making overly confident predictions is strongly penalized.  This is evidenced by the low KL-divergence \wrt to underlying data distribution that our objective achieves (see \figref{fig:sample_plot}).
\item \textbf{Improves sample efficiency.}
    As shown in \figref{fig:sample_plot}, even with fewer samples as compared to Maximum Likelihood training (with and without L2 regularization), our model is able to more accurately model the true data-distribution as evidenced by the significantly lower KL divergence \wrt to the data-generating model.
\end{compactenum}
Further details are provided in the supplement.

\subsection{Attribute Prediction.}
\label{sec:mlml}
\textbf{Datasets and Models.} We now replicate the synthetic setup in a multi-label image attribute prediction setting on two real world datasets -- Animals with Attributes (AWA; \citet{xian2017zero}) and Caltech UCSD Birds 200-2011 (CUB; \citet{wah2011caltech}).
Specifically, we randomly sub-sample from the set of all positive attributes for an image and evaluate the performance of models based on their ability to recover all annotated attributes for each image.
%

\begin{table}
\centering
\resizebox{0.95\columnwidth}{!}{
\setlength{\tabcolsep}{10pt} 
\renewcommand{\arraystretch}{1}
\begin{tabular}{c  c c  c c c c}\toprule
    &\textbf{Method} & \textbf{\# observed} & \multicolumn{3}{c}{\textbf{Average Precision@k}} \\
    \cmidrule(lr){4-6}
    && \textbf{labels} & \texttt{@1} & \texttt{@5} & \texttt{@10} \\  
    \midrule
    \multirow{4}{*}{\rotatebox{90}{AWA}}&CE+L2 &  \multirow{2}{*}{1}     & 46.82 & 51.03 & 54.18\\
    &Ours &     & \textbf{49.23} & \textbf{53.46} & \textbf{57.12}\\
    \cmidrule(lr){2-6}
    &CE+L2 & \multirow{2}{*}{20\%}  & 52.74 & 57.48 & 63.71   \\  
    &Ours &  & \textbf{56.79} & \textbf{61.54} & \textbf{66.28}   \\  
    \midrule
    \multirow{4}{*}{\rotatebox{90}{CUB}}&CE+L2 & \multirow{2}{*}{1}     & 27.10 & 31.40 & 35.62\\
    &Ours &    & \textbf{29.32} & \textbf{33.19} & \textbf{38.83}\\
    \cmidrule(lr){2-6}
    &CE+L2 & \multirow{2}{*}{20\%}  & 32.42 & 35.94 & 39.21\\
    &Ours &  & \textbf{35.64} & \textbf{38.20} & \textbf{43.01} \\
\bottomrule
\end{tabular}}\\[-5pt]
\caption{In the multi-label classification with missing labels setting, we observe that our proposed method outperforms standard cross-entropy training with $\sim3\%$ improvements when k=10 on both AWA and CUB datasets with both just one randomly sampled label and observing $20\%$ of the annotations}
\label{tab:mlml}
\vspace{-10pt}
\end{table}
While AWA contains ${\sim}30K$ images across 50 categories and 85 attributes, CUB is much sparser with ${\sim}11K$ images across 200 categories and 312 attributes.
We report results under two aggressive missing-label settings; sampling either only a single attribute or 20\% of the annotated attributes for each image. 
For both L2 regularized cross-entropy (CE+L2) and our loss, we use the pre-final layer activations of Resnet-152 \cite{he_cvpr16} as the image-representation and train a two-layered MLP, optimized using Adam \cite{kingma_arxiv14} with a learning rate of $1e{-}5$ and a batch size of $64$. 
Early-stopping was used to pick the best set of parameters. 

\textbf{Results.} Unlike the toy setting in \secref{sec:toy_exp}, the data generating distribution is unknown and so, we cannot evaluate using KL-divergence \wrt it. 
As is common to the multi-class with missing labels setting, we use \texttt{Average Precision@k} -- that measures the number of correct annotations present in the top-$k$ ranked predictions.
From \tabref{tab:mlml} it can be gleaned that our approach outperforms standard cross-entropy training in both settings (just one randomly sampled label and $20\%$ of the annotated labels) and on both AWA and the much harder, CUB datasets.
For instance, our method achieves an improvement of $\sim3\%$ when evaluated on the top-10 retrieved predictions.

\subsection{Visually Grounded Language Generation}
\label{sec:cap}
\noindent\textbf{Datasets.} We report results on standard image-captioning datasets -- Flickr-8k \cite{hodosh_jair13}, Flickr-30k \cite{young_acl14} and COCO \cite{coco}.
We use standard splits \cite{karpathy_cvpr15} of size $1000$ to report results on the first two and a test split of size $5000$ for the COCO dataset.
Futher, to mimic the problem of missing labels we train only on a single (arbitrarily chosen) caption while evaluating on all 5 captions. 
This approach helps us to evaluate if our model learns to place beliefs on other unseen but `correct' outputs while seeing only \emph{uni-modal} training data. 
Owing to the small size of the PASCAL-50S dataset \cite{jas_cvpr15}, we only evaluate on it while using a model trained on COCO.

For VQG, We use three datasets built on a small subset of COCO, Flickr and Bing images \cite{mostafazadeh_arxiv16}.
We train on ${\sim}2.5K$ images and report results on a test of size ${\sim}1.5K$ for each dataset. 
To stay consistent with the captioning experients, we report retrieval numbers on a randomly chosen subset of 1000 images and their 5 corresponding questions from the test set.
\\ \\
\noindent \textbf{Models.}
For both tasks, we train a model similar to \citet{lu_cvpr17} that uses activations from an ImageNet pre-trained Resnet-152 \cite{he_cvpr16} architecture as image-representations.  
For both captioning and VQG, the learnt LSTM model has one layer, 1024-dimensional hidden states and is optimized using Adam \cite{kingma_arxiv14} with a learning rate of $1e{-}4$.
The similarities $\calK_{ij}$ used to weigh the supervision from neighboring data-points are computed in a learnt space got by projecting the image-representations through a 2-layered MLP with 512 hidden units in each layer. 
As discussed in sec.~\ref{sec:implementation_details}, the learning rate for this transformation is is 10$\times$ smaller compared to the LSTM parameters. 
\begin{table}[t]
\centering
\resizebox{0.95\columnwidth}{!}{
\begin{tabular}{ c c  c c c c c@{}}\toprule
     & \multirow{2}{*}{\textbf{Method}} & \multicolumn{3}{c}{\textbf{Oracle Metrics @20}} & \multirow{2}{*}{\textbf{\vtop{\hbox{\strut ~distinct}\hbox{\strut 4-grams}}}} & \multirow{2}{*}{\textbf{\vtop{\hbox{\strut Recall\textsubscript{5}}\hbox{\strut @100}}}} \\
     \cmidrule(lr){3-5}
     &  & \textbf{CIDEr} & \textbf{SPICE} & \textbf{METEOR} & & \\
     \midrule
    \multirow{5}{*}{\rotatebox{90}{Flickr-8k}} & MLE & 0.5072 & 0.1564 & 0.1553 & 2205 & 0.71\\
    & \cite{rennie_arxiv16}            & \textbf{0.5272} & 0.1509 & 0.1498 & 1834 & 0.74\\
     & \cite{dai_arxiv17}         & 0.4982 & 0.1598 & 0.1420 & 1730 & 0.72\\
     & Caption-Transfer-Without-Attention                          & 0.5181 & 0.1561 & 0.1565 & \textbf{2503} & 0.89\\
     & Caption-Transfer-With-Attention                   & 0.5240 & \textbf{0.1620} & \textbf{0.1614} & 2498& \textbf{0.95} \\
    \midrule
    \multirow{5}{*}{\rotatebox{90}{Flickr-30k}} & MLE & 0.6729 & 0.1642 & 0.1723 & 1920 & 1.30 \\
    & \cite{rennie_arxiv16}             & 0.6832 & 0.1520 & 0.1689 & 1824 & 1.36 \\
    & \cite{dai_arxiv17}          & 0.7120 & 0.1692 & 0.1752 & 1730 & 1.34 \\
    & Caption-Transfer-Without-Attention                           & 0.7180 & 0.1721 & 0.1794 & 1822 & 1.52 \\
    & Caption-Transfer-With-Attention                    & \textbf{0.7246} & \textbf{0.1802} & \textbf{0.1843} & \textbf{2101} & \textbf{1.63} \\
    \midrule
    \multirow{8}{*}{\rotatebox{90}{COCO}}    & MLE & 0.8014 & 0.2132 & 0.2245 & 4218 & 1.45 \\
    \cmidrule(lr){2-7}
    & augment & 0.8294 & 0.2147 & 0.2331 & 3766 & 1.49 \\
    & no-refine & 0.8316 & 0.2182 & 0.2304 & 4117 & 1.63 \\
    \cmidrule(lr){2-7}
    & \cite{rennie_arxiv16} & 0.8410 & 0.2013 & 0.2272 & 3988 & 1.53 \\
    & \cite{dai_arxiv17} & 0.8120 & 0.2117 & 0.2340 & 4011 & 1.49 \\
    & Caption-Transfer-Without-Attention & 0.8398 & 0.2165 & 0.2378 & 4128 & 1.78 \\
    & Caption-Transfer-With-Attention  & \textbf{0.8422} & \textbf{0.2210} & \textbf{0.2405} & \textbf{4270} & \textbf{1.84} \\
\bottomrule
\end{tabular}}\\[-5pt]
\caption{While reporting standard task-specific and diversity metrics, we observe that our methods outperform standard MLE and the baselines on all three datasets Flickr-8k, Flickr-30k and COCO on the retrieval task that measures multi-modal output mappings. 
Further, note that the task-specific metrics like CIDEr and SPICE are generally lower since we train using only one caption.}
\label{tab:cap}
\vspace{-10pt}
\end{table} 

\noindent\textbf{Ablations}
Recall that in \secref{subsec:fac_loss} we described two variants of our approach for sequence prediction:\\[-15pt] 
\begin{compactenum}[\hspace{3pt}1.]
\item \textbf{Caption-Transfer-without-Attention}: where we transfer entire captions from  neighboring images, and
\item \textbf{Caption-Transfer-with-Attention}: where we use attention models to selectively weigh relevant portions of the neighbor's caption. 
\end{compactenum}

\noindent\textbf{Baselines.}
In addition to standard maximum likelihood training we also compare to two ablations of our method:  
\begin{compactenum}
\item \textbf{augment} -- captions of neighboring images are directly appended as ground truth to create a larger training set (corresponds to setting $\calK_{ij}=1$).
    Improvements on this setting indicate the advantage of both softly-enforcing the neighborhood as well as learning the representation space for computing similarities. 
\item \textbf{no-refine} -- Unlike the full setting, the representation is held fixed and is not refined from the generic representations to specialize for the task at hand. 
    Improvements over this baseline denotes the advantages of jointly learning the representation space (and adapting the neighborhood) apart from transfering supervision.
\end{compactenum}
Further, we compare to two other strong methods that do not employ MLE --- \cite{dai_arxiv17} use adversarial training to distinguish between human and generated captions and \cite{rennie_arxiv16} directly optimize for a task-specific metric using policy gradients with a novel variance reduction baseline.
\\ \\
\noindent\textbf{Evaluation Metrics.} We evaluate the sequence generation models using the following metrics that each evaluate for certain desirable properties of the model --
\begin{compactenum}[\hspace{0pt} 1.]
    \item \emph{Oracle Metrics.} Each output is evaluated against the reference sequences using standard captioning metrics like CIDER or SPICE. 
        Following previous works \cite{guzman2014efficiently, lee_nips16, snellstochastic} that consider ambiguous tasks, we evaluate the decoded lists using \emph{oracle} metrics that report the best output in the list -- mimicking an `oracle' user that selects the most suitable option for a downstream task.
    \item \emph{Diversity Metrics.} Apart from producing high-quality captions, linguistic diversity in the decoded lists provides a good signal for the multi-modal nature of the learnt model. 
        Similar to \cite{li_arxiv15}, we measure diversity in the decoded lists by reporting the number of distinct $n$-grams (normalized by sentence length) present in the decoded lists. 
    \item \emph{Retrieval Metrics.} A drawback of the first two metrics is that they also depend on the inference procedure used to decode the output lists (\eg beam search). 
        Therefore, we directly evaluate the beliefs placed by the model on different outputs in a retrieval setting where a pool of human-annotations are ranked based on their $\log$-probability under the model for a given image.
    Then, we compute \texttt{Recall\textsubscript{m}@k} that evaluates for -- \emph{the average number of the $m$ ground truth captions that were present in the top-k retrieved sequences}. 
\end{compactenum}

\textbf{Results.} For both captioning and VQG, we decode output lists for all methods using beam search with a beam size of 20. 
As can be seen from \tabref{tab:cap} and \tabref{tab:vqg}, both variants of our approach outperform cross-entropy training on all three standard image-captioning metrics. 
Excluding Flickr-30K for captioning and Bing for VQG, our approach performs the best in terms of output quality (as evidenced by higher oracle numbers).
Further, our approach achives the best performance on both diversity and retrieval metrics indicative of the multi-modal mapping learnt by the model on both tasks of interest.
Additionally, our methods outperform both \emph{hard} and \emph{no-refine} ablations of our method -- we only show the performance for captioning on COCO owing to space constraints and provide the rest in the supplement.

\begin{table}
\centering
\resizebox{0.95\columnwidth}{!}{
\begin{tabular}{ c c  c c c c c@{}}\toprule
     & \multirow{2}{*}{\textbf{Method}} & \multicolumn{3}{c}{\textbf{Oracle Metrics @20}} & \multirow{2}{*}{\textbf{\vtop{\hbox{\strut ~distinct}\hbox{\strut 4-grams}}}} & \multirow{2}{*}{\textbf{\vtop{\hbox{\strut Recall\textsubscript{5}}\hbox{\strut @100}}}}\\ 
     \cmidrule(lr){3-5}
     &  & \textbf{CIDEr} & \textbf{SPICE} & \textbf{METEOR} & & \\
     \midrule
    \multirow{5}{*}{\rotatebox{90}{Flickr}} & MLE & 0.3510 & 0.1201 & 0.1273 & 1294 & 0.35\\
    & \cite{rennie_arxiv16} & 0.3822& 0.1240& 0.1298& 1350 & 0.38\\
    & \cite{dai_arxiv17} & 0.3572 & 0.1286 & 0.1287 & 1258 & 0.34\\
    & Caption-Transfer-Without-Attention & 0.3720 & 0.1292 & 0.1392 & 1388 & 0.46\\
     & Caption-Transfer-With-Attention & \textbf{0.3827} & \textbf{0.1445} & \textbf{0.1462} & \textbf{1395} & \textbf{0.51} \\
    \midrule
    \multirow{5}{*}{\rotatebox{90}{COCO}} & MLE & 0.3233 & 0.1182 & 0.1245 & 967& 0.33 \\
    & \cite{rennie_arxiv16} & 0.3485 & 0.1224 & 0.1209 & 1104& 0.35 \\
    & \cite{dai_arxiv17} & 0.3296 & 0.1215 & 0.1241 & \textbf{1270}& 0.38 \\
    & Caption-Transfer-Without-Attention & 0.3471 & 0.1262 & 0.1276 & 1192& 0.43 \\
    & Caption-Transfer-With-Attention & \textbf{0.3506} & \textbf{0.1282} & \textbf{0.1304} & 1220& \textbf{0.49} \\
    \midrule
    \multirow{5}{*}{\rotatebox{90}{Bing}} & MLE & 0.2890 & 0.1202 & 0.1309 & 755& 0.37 \\
    & \cite{rennie_arxiv16} & \textbf{0.3681} & 0.1225 & 0.1287 & 910& 0.38 \\
    & \cite{dai_arxiv17} & 0.3224 & 0.1199 & 0.1256 & 937& 0.42 \\
    & Caption-Transfer-Without-Attention & 0.3355 & 0.1289 & 0.1296 & 984& 0.49 \\
    & Caption-Transfer-With-Attention & 0.3410 & \textbf{0.1314} & \textbf{0.1336} & \textbf{1021} & \textbf{0.55} \\
\bottomrule
\end{tabular}}\\[0pt]
\caption{We find that our methods outperform the baselines on all three datasets Flickr, COCO and Bing datasets on the retrieval task that evaluates ability of the model to make multi-modal output maps.
Similar to image-captioning, note that the task-specific metrics are generally lower as we train using only one question.}
\label{tab:vqg}
\vspace{-10pt}
\end{table}

\section{Discussion} \label{sec:disc}

\begin{figure}
\centering
\includegraphics[width=4cm,height=3cm]{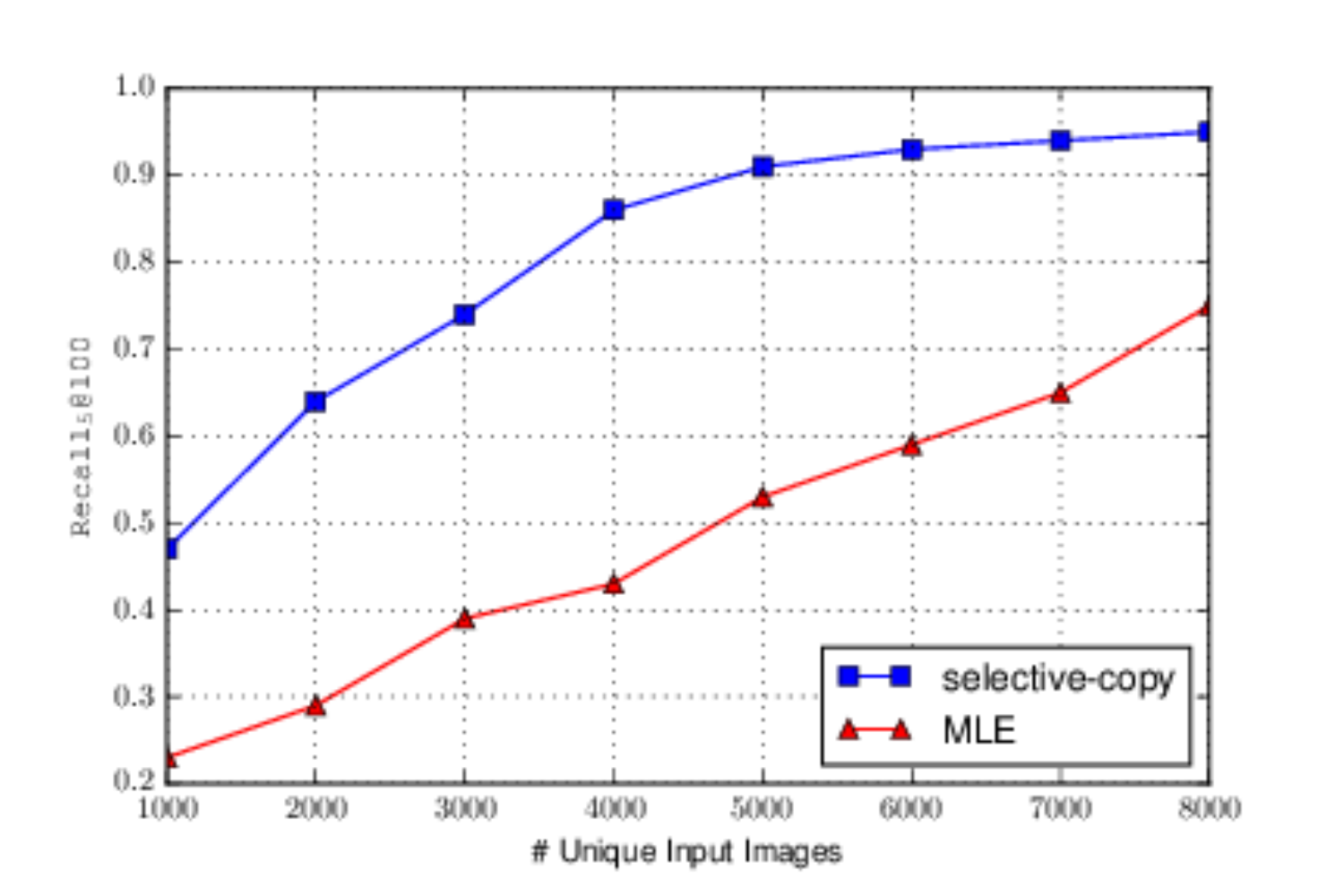}
\includegraphics[width=4cm,height=3cm]{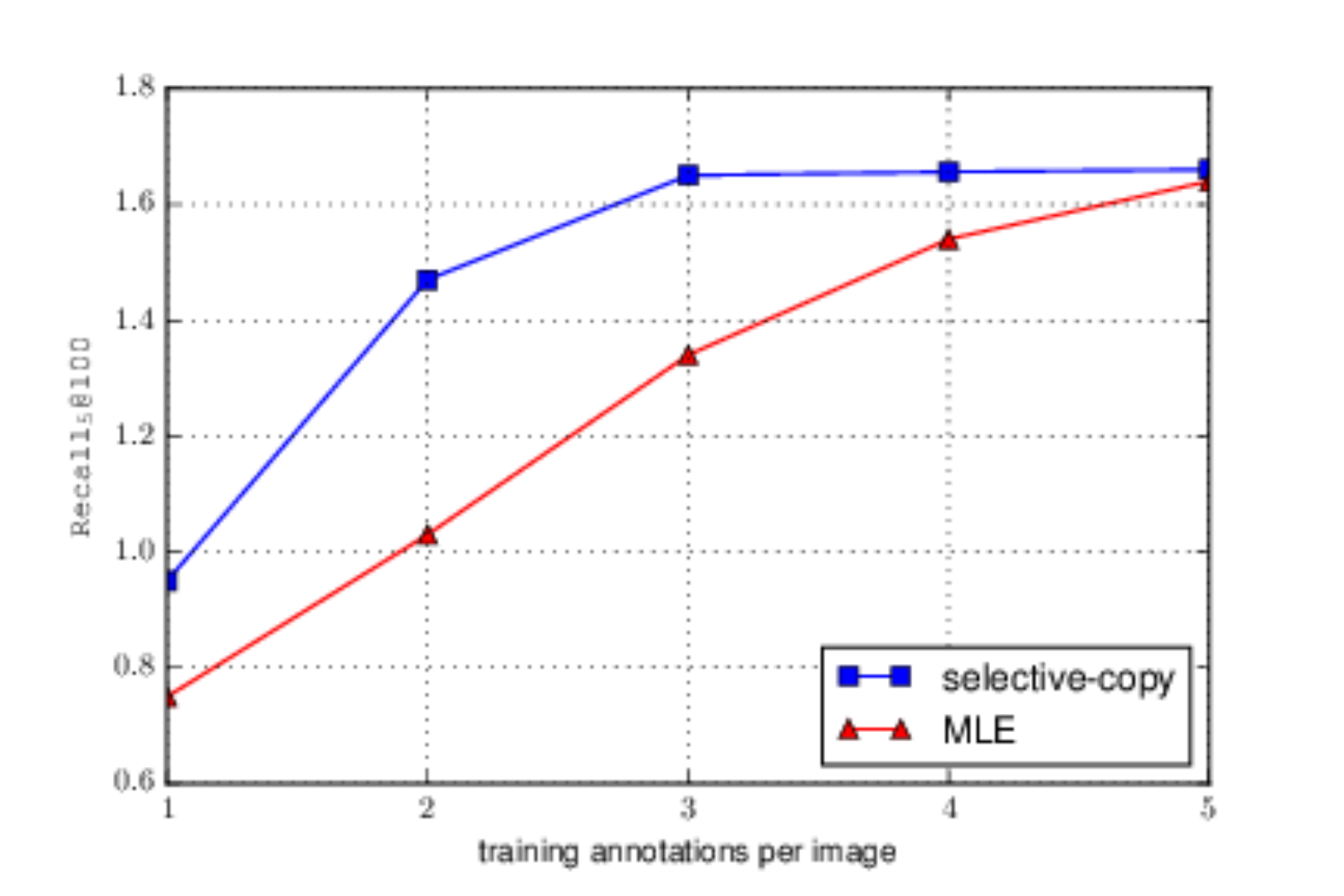}\\[-8pt]
\caption{Our method is sample efficient compared to standard cross-entropy training. On the Flickr8k image-captioning task, our method performs comparably to CE with the full dataset  while only using ${\sim}3K$ images \textbf{(left)}. Similar trends exist as the number of annotations per image is increased \textbf{(right)}.}
\label{fig:sample}
\vspace{-10pt}
\end{figure}


\textbf{Sample Efficiency on Captioning.} We perform sample-efficiency experiments similar to those in the toy-setting for captioning on the Flickr-8k dataset in two ways --
1) by gradually increasing the number of unique images used from 1000 to 8000 while still using only one caption annotation. 2) by gradually increasing the number of additional ground truth captions used from 1 to 5. 
As seen in \figref{fig:sample} (left), our approach obtains a Recall\textsubscript{5}@100 score of 1.58 using only $5K$ images which is very close to what is obtained using all $8k$ images (=1.60) in the first case.
In the second case, we observe in \figref{fig:sample} (right) that using only 3 captions per image (1.65) nearly obtains the same retrieval score as using all the 5 captions (=1.66).
This demonstrates that our approach can lead to efficient learning of the true distribution even in data-sparse regimes.
\\ \\
While the primary focus of our work is to learn multi-modal mappings even with access to uni-modal datasets, we observe that our proposed method performs competitively when trained and evaluated under standard captioning settings \ie all 5 captions are used during training and \emph{one} best output is evaluated (as against using oracle metrics).
For instance, our method achieves a METEOR and CIDEr score of 0.28 and 1.14 respectively, slightly outperforming \citet{lu_cvpr17} (0.27 and 1.09) on the COCO-captioning task.
We observe similar trends on question-generation and include detailed results in the supplement.

\section{Conclusion}
\noindent In this work, we propose a novel objective that incorporates the inductive bias that the outputs of neighboring data points can be used to provide additional supervision especially when obtaining exhaustive annotations is expensive or worse, intractable. 
The proposed objective allows the model to place beliefs on multiple plausible outputs while still observing only one annotation per input.
We first study the properties of our method on a synthetic dataset where the underlying data-distribution is known allowing us to control the difficulty of the experiments and directly evaluate the learnt posteriors.
Further, we replicate this toy setting on a real-world multi-label prediction problem using standard attribute datasets. 
Finally, we show that our approach leads to better quality outputs with higher diversity on two well-established visually grounded language-generation tasks -- captioning and question generation. 
We observe that our approach outperforms various ablations and baselines on both tasks on the various evaluation metrics used.
\bibliography{strings,ashwinkv}
\bibliographystyle{icml2018}





\end{document}